\documentclass[letterpaper, 10 pt, conference]{ieeeconf}  

\IEEEoverridecommandlockouts                              

\usepackage{cite}
\usepackage{multirow}
\usepackage{graphicx} 
\usepackage{booktabs}
\usepackage[namelimits]{amsmath} 
\usepackage{amssymb}             
\usepackage{amsfonts}           
\usepackage{mathrsfs}           
\usepackage{lipsum}
\usepackage{url}
\usepackage{algorithm}
\usepackage{algpseudocode}
\usepackage{hyperref}
\usepackage[capitalize]{cleveref}
\hypersetup{
    colorlinks=true,
    linkcolor=blue,
    filecolor=magenta,      
    urlcolor=cyan
}

\title{\LARGE \bf
DTPP: Differentiable Joint Conditional Prediction and Cost Evaluation for Tree Policy Planning in Autonomous Driving
}

\author{Zhiyu Huang$^1$, Peter Karkus$^2$, Boris Ivanovic$^2$, Yuxiao Chen$^2$, Marco Pavone$^{2,3}$, and Chen Lv$^1$ 
\thanks{$^1$School of Mechanical and Aerospace Engineering, Nanyang Technological University, Singapore. {\tt zhiyu001@e.ntu.edu.sg, lyuchen@ntu.edu.sg}}%
\thanks{$^{2}$NVIDIA Research, NVIDIA Corporation, 
Santa Clara, CA, USA. {\tt \{pkarkus, bivanovic, yuxiaoc, mpavone\}@nvidia.com
}}%
\thanks{$^3$Department of Aeronautics and Astronautics, Stanford University.}
}

\begin{document}

\maketitle
\thispagestyle{empty}
\pagestyle{empty}

\begin{abstract}

Motion prediction and cost evaluation are vital components in the decision-making system of autonomous vehicles. However, existing methods often ignore the importance of cost learning and treat them as separate modules. In this study, we employ a tree-structured policy planner and propose a differentiable joint training framework for both ego-conditioned prediction and cost models, resulting in a direct improvement of the final planning performance. For conditional prediction, we introduce a query-centric Transformer model that performs efficient ego-conditioned motion prediction. For planning cost, we propose a learnable context-aware cost function with latent interaction features, facilitating differentiable joint learning. We validate our proposed approach using the real-world nuPlan dataset and its associated planning test platform. Our framework not only matches state-of-the-art planning methods but outperforms other learning-based methods in planning quality, while operating more efficiently in terms of runtime. We show that joint training delivers significantly better performance than separate training of the two modules. Additionally, we find that tree-structured policy planning outperforms the conventional single-stage planning approach. Code is available: \url{https://github.com/MCZhi/DTPP}.

\end{abstract}

\section{INTRODUCTION}

A fundamental requirement for autonomous vehicles is the ability to make decisions that are safe, informed, and human-like. Achieving this involves accurate prediction of the future behavior of traffic participants and planning that ensures safety, comfort, and adherence to traffic norms. 
Due to inherent uncertainties in the real world, the decision-making system should be capable of \emph{policy planning} that accounts for different futures and options for the ego vehicle to react. 
To tackle this intractable continuous-space planning problem, tree-structured policy planners, such as Tree Policy Planning (TPP) \cite{chen2023tree} and Monte-Carlo Tree Search (MCTS) \cite{li2022efficient}, employ a tree policy where the optimal action can be found by solving a discrete Markov Decision Process (MDP). Specifically, the TPP algorithm constructs two trees: a trajectory tree and a scenario tree. Each branch of the trajectory tree represents a candidate (multi-stage) trajectory for the ego agent, and each branch of the scenario tree contains the predicted outcomes of neighboring agents.

\begin{figure}
    \centering
    \includegraphics[width=\linewidth]{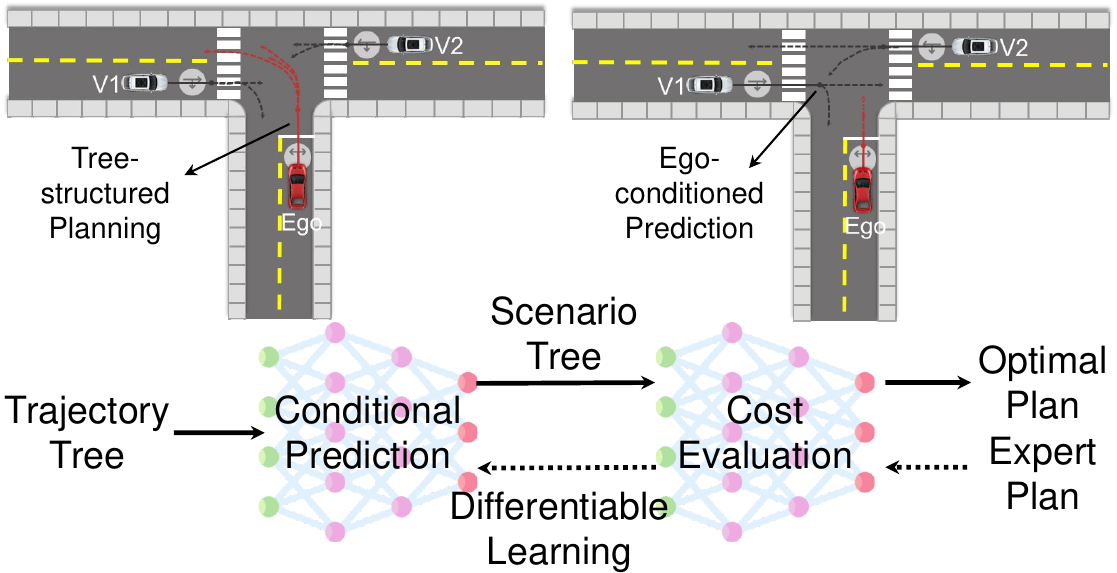}
    \caption{Overview of the proposed decision-making framework. A tree-structured planner generates a multi-stage trajectory tree, which is then used by the conditional prediction model to generate a scenario tree. The optimal plan is selected using the cost evaluation of both trees, and the conditional prediction and cost evaluation models are jointly learnable.}
    \label{fig:0}
    \vspace{-0.4cm}
\end{figure}

However, tree-structured planners face two significant challenges. First, in contrast to most neural motion prediction models that only predict unconditional future trajectories of other agents (i.e., without considering bi-directional interactions), tree-structured planners require a prediction model capable of efficiently producing ego-conditioned predictions. The prediction model must be able to handle varying search depths (timesteps) within the planning process, maintain causal relationships \cite{chen2023tree}, and efficiently process multiple branches.
Second, evaluating the cost (or reward) of actions in alignment with human decision-making is challenging. Existing approaches often employ a simple linear cost function with a set of manually-crafted features and fixed weights~\cite{huang2021driving, wu2020efficient}. However, factors related to agent interactions are intricate to design and difficult to quantify manually. Accordingly, a learnable cost function becomes an attractive approach to reflect human driving preferences.

Our key contributions are threefold. First, we present the DTPP framework that integrates efficient ego-conditioned predictions with the learning of a situation-aware cost function (\cref{fig:0}). Our framework can flexibly leverage both learned and handcrafted cost components, is end-to-end differentiable, and enables joint training with human driving data.
Second, we present a novel query-centric, Transformer-based prediction model that enables efficient multi-stage motion predictions conditioned on multiple potential future ego trajectories.
The key advantage of our prediction model is its use of the ego agent's future branch information (trajectory tree) as a query in its Transformer decoder, in contrast to the conventional approach of incorporating a single ego trajectory plan during context encoding~\cite{huang2023conditional, tolstaya2021identifying, song2020pip, salzmann2020trajectron++, ngiam2021scene}. Finally, we demonstrate that this approach yields strong performance for both closed-loop planning and prediction across a wide array of challenging, interactive scenarios in the large-scale, real-world nuPlan dataset~\cite{nuplan}.

\section{Related Work}

{\bf Conditional Motion Prediction.}
There has been significant progress in the development of motion prediction models based on neural networks \cite{huang2022multi, mo2022multi, shi2022motion, huang2023gameformer, jiang2023motiondiffuser}, including recent query-centric Transformer models~\cite{zhou2023query, shi2023mtr++}.
However, the majority of these prediction models overlook the influence of the controllable agent (ego) on the behavior of other agents. To address this, the concept of (ego-)conditional motion prediction (CMP) has been introduced. Most existing research on CMP simply incorporates the planned ego trajectory in the encoding stage as part of the environment context \cite{huang2023conditional, song2020pip, tolstaya2021identifying, salzmann2020trajectron++}. These models are computationally inefficient when used for closed-loop policy planning as predictions need to be repeatedly computed for multiple ego trajectories. 
Some approaches incorporate ego information in a later, decoding stage, e.g., within a RNN cell \cite{huang2023learning, tang2022interventional}, or by employing masked attention in a Transformer model \cite{ngiam2021scene, yuan2021agent}. While these models are more efficient for repeated ego-conditioned predictions, they only take one ego trajectory input at a time. In contrast to prior work, we propose a query-centric Transformer model for CMP that incorporates an entire ego \emph{trajectory tree} in its query vector, thus significantly reducing runtime. Additionally, we train our CMP model jointly with the cost model, providing additional regularization for conditioning on ego plans other than the ground truth and directly optimizing the model to be effective for planning.

{\bf Learning Cost Functions.}
Designing a cost function for autonomous vehicle motion planning is a well-known challenge, partly due to the difficulty of expressing and combining human preferences, safety, efficiency, and comfort. As a result, many prior works propose learning cost functions from expert data, e.g., using inverse reinforcement learning (IRL)~\cite{arora2021survey}. In particular, maximum entropy IRL has been applied to model driver behavior~\cite{kuderer2015learning, sun2020expressing, huang2021driving} or planning tasks~\cite{rosbach2020planning, wu2020efficient, huang2023conditional} by learning the weights of a linear cost function. However, a linear cost function poses overly-strong assumptions, resulting in suboptimal or inaccurate evaluations in some scenarios \cite{huang2023conditional}. Moreover, manually-designed cost features may not capture the nuanced decision-making patterns of human drivers. To address these limitations, Deep IRL \cite{wulfmeier2017large} directly learns cost maps using neural networks without relying on a linear cost structure, albeit with a substantial computational burden. Inspired by \cite{phan2023driveirl}, we propose learning a set of context-aware cost features as well as a cost weight decoder that weights learned and hand-crafted cost features. We then train the cost model jointly with the CMP model. 

{\bf Joint Prediction and Planning}
Joint prediction and planning are necessary when dealing with highly-interactive driving scenarios \cite{hagedorn2023rethinking}. Several studies adopt a holistic neural network approach to jointly generate trajectories for both the ego agent and other involved agents \cite{scheel2022urban, vitelli2022safetynet, huang2023gameformer}. However, these planners lack reliability as they fail to explicitly consider the outcomes or costs of the actions of the ego agent. \cite{ivanovic2020mats} presents a planning framework that utilizes the prediction results as control system parameters, which significantly improves computational efficiency. However, its performance is limited due to the simple affine structure of its prediction representation. Some works advocate for differentiable joint prediction and planning structures \cite{huang2023differentiable, karkus2023diffstack}, rendering the prediction results more amenable to downstream planning and enabling learning cost functions from data. However, these methods cannot model the bi-directional interaction between the ego and other agents. Therefore, we integrate a tree-structured planner with efficient CMP and neural cost evaluation models to enable joint training of prediction and planning modules.

\section{Methodology}


\begin{figure*}
  \begin{minipage}[b]{.65\textwidth}
     \centering
     \includegraphics[width=0.96\textwidth]{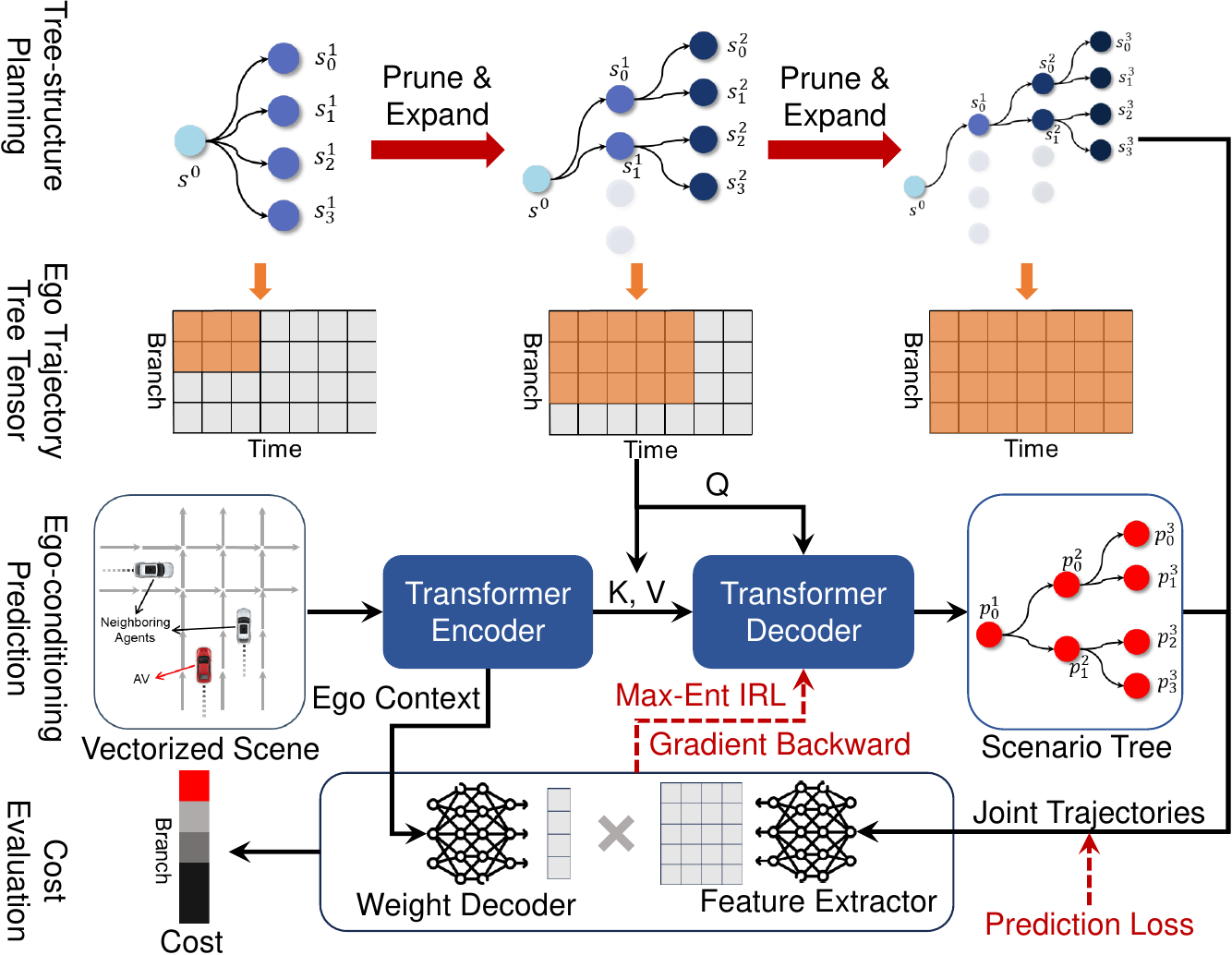}
     \caption{Overview of the DTPP framework with joint learnable prediction and cost evaluation models. The framework encompasses iterative node pruning and expansion, guided by ego-conditioned prediction outcomes and cost evaluations. During training, the loss of cost evaluation can be back-propagated to the prediction module, enabling differentiable and joint optimization of both modules.}
     \label{fig:1}
  \end{minipage}
  \
  \begin{minipage}[b]{.32\textwidth}
    \centering
    \begin{algorithm}[H]
    \small
    \caption{Tree Policy Planning with Learned Prediction and Cost Models}
    \begin{algorithmic}[1]
    \Require $N_l$ expansion stages, $f_p$ prediction model, $f_c$ cost model, $f_e$ node expansion function, $s^0$ initial node.
    \State Encode scene context using $f_p$ (encoder)
    \For{$i \gets 0$ to $N_l-1$}
    \State Expand the current node(s) $s^i$ using function $f_e$ to obtain the trajectory tree $s^{i+1}$
    \State Query the prediction model $f_p$ (decoder) with the trajectory tree to obtain the scenario tree $p^{i+1}$
    \State Query the cost model $f_c$ with the trajectory tree and scenario tree to obtain branch costs $c^{i+1}$
    \State Prune nodes in $s^{i+1}$ using $c^{i+1}$
    \EndFor
    \State Compute the optimal first-stage node $s^{1*}$ using dynamic programming
    \end{algorithmic}
    \label{algo1}
    \end{algorithm}
  \end{minipage}%
  \vspace{-0.4cm}
\end{figure*}

The proposed DTPP framework with its associated prediction and cost models are illustrated in \cref{fig:1} and \cref{algo1}. Our planner is based on top of TPP~\cite{chen2023tree}, with the important additions of node pruning enabled by our learned cost model, and integration with our CMP model, which together significantly improve planning performance and efficiency.

The key idea behind tree-structured planning is to approximate the intractable continuous-space policy planning problem by sampling a discrete set of ego trajectories in multiple stages, forming a \emph{trajectory tree}, and predicting the motion of other agents conditioned on each ego trajectory segment, forming a \emph{scenario tree}. The optimal ego action is then derived using dynamic programming. 
Specifically, trajectory tree nodes $s_{i}^{j}$ contain an ego trajectory spanning from $t_{0}^{j}$ to $t_{F}^{j}$, where $t_{0}^{j}$ and $t_{F}^{j}$ indicate the start and end times of the $j$-th stage ($j \ge 1$).
Scenario tree nodes $p_{i}^{j}$ contain the state of the environment (predicted states of nearby agents), and we use the same labeling convention. To build the trees, we iteratively sample a set of target states for each $s_{i}^{j}$ and generate trajectories that connect $s_{i}^{j}$ to the respective target states. We then predict environment states $p_{i}^{j}$  with a CMP model. We compute costs with the evaluation model and prune low-scoring nodes. We continue to expand the trees until a specified time limit or depth is reached. 

Our framework has two key novel model components. First, the query-centric CMP model extracts an environment context with a Transformer encoder, and outputs joint trajectories of other agents (scenario tree) conditioned on possible ego plans with a Transformer decoder. Importantly, the ego trajectories enter the model as part of the decoder query vector, so during planning the encoder is only called once and the decoder is only called once per planning stage.
Second, the cost evaluation module takes the trajectory tree and scenario tree as inputs to compute the costs of each branch (plan). To do this, it combines a neural feature extractor (from the joint trajectories of all agents) and a neural weight decoder (from the encoded ego context). The cost model is learned jointly in the context of the planner and prediction model, so it is optimized to capture the desired driving behavior and act as an effective pruning function in testing.

\subsection{Planning Process}
\label{sec:planning}
Our proposed planning algorithm is outlined in \cref{algo1}. 
The tree policy planner repeatedly expands the selected nodes guided by the learned prediction and cost models until reaching the maximum expansion stage $N_l$. 
By introducing cost learning as an effective heuristic for expansion, we can ensure that we only expand promising nodes for efficiency. We then use dynamic programming to find the optimal initial stage node for execution \cite{chen2023tree}. 
While our framework is extensible to any number of stages, we opt for $N_l=2$ in experiments. 


To construct the trajectory tree, we use a lane-centric target sampling process that incorporates mapping and routing information. First, we find multiple reference paths (e.g., lane centerlines) for the ego vehicle \cite{huang2023gameformer}. Then, we sample varying target speeds and generate candidate trajectories across different paths based on the target speeds. Note that this is different from TPP, which directly samples target states, because this offers more stable planning performance in various scenarios with complex road networks. Given the current state of the ego vehicle at a node $s(0) = [x_0, y_0, v_0, a_0, \theta_0, l_0]$, representing its coordinates, velocity, acceleration, heading, and position on a reference path, we adopt a third-order polynomial to parameterize the trajectory's velocity $v(t)$ concerning the target speed $v_{\text{target}}$:
\begin{equation}
\begin{split}
v(t) &= a_0 + a_1 t + a_2 t^2 + a_3 t^3, \\
v(0) &= v_0, \ v(t_F) = v_{\text{target}}, \\
a(0) &= \dot v(0) = a_0, \ a(t_F) = \dot v(t_F) = 0.
\end{split}
\end{equation}
We can directly compute the coefficients of $v(t)$ in closed form and determine the velocity of the trajectory. Then, we can calculate the positions on the reference path $l(t) = l_0 + \int_{0}^{t} v(t) dt $ and the coordinates $x(t)$, $y(t)$, $\theta(t)$ in Cartesian space. Finally, we drop trajectories that violate dynamic constraints. 
We set an upper limit on the number of nodes within the trajectory tree and randomly discard nodes if the limit is exceeded.

\subsection{Conditional Prediction Model} 
\textbf{Transformer encoder}.
The inputs to the model are the past trajectories of  $N_a = 10$ nearby agents and the ego vehicle, as well as $N_m = 50$ nearby map elements including lanes and crosswalks. Agent trajectories are $T_h=20$ timesteps long with $\Delta T=0.1s$ resolution. Trajectories are encoded with LSTM networks (with shared weights); map elements are encoded with MLP networks. The encoding vectors are concatenated to form an environment context with shape $[1+N_a+N_m, \ D]$, $D=256$.
To capture inter-dependencies among scene elements, we process the environment context using a self-attention Transformer encoder consisting of $L_e = 3$ layers, and output dimension $[1+N_a+N_m, \ D]$.
Importantly, this encoder is run only once during the planning process.

\textbf{Transformer decoder}. 
The decoder is designed to efficiently extract information from both the environment context and ego trajectory tree, generating multi-stage predictions for the entire scenario tree in a single forward pass. Ego-conditioning is achieved through two aspects: the query to the attention modules containing the ego's future information, and the ego trajectory tree as information to the attention module. Initially, we encode the ego trajectory tree through an MLP, yielding the ego trajectory tree tensor. To handle the variable-sized trajectory tree, we organize the branches into a fixed-size tensor with a maximum branch size of $M=30$ and a maximum time step of $T=8 \ s$, padding invalid positions with zeros. 

\begin{figure}[t]
    \centering
    \includegraphics[width=0.93\linewidth]{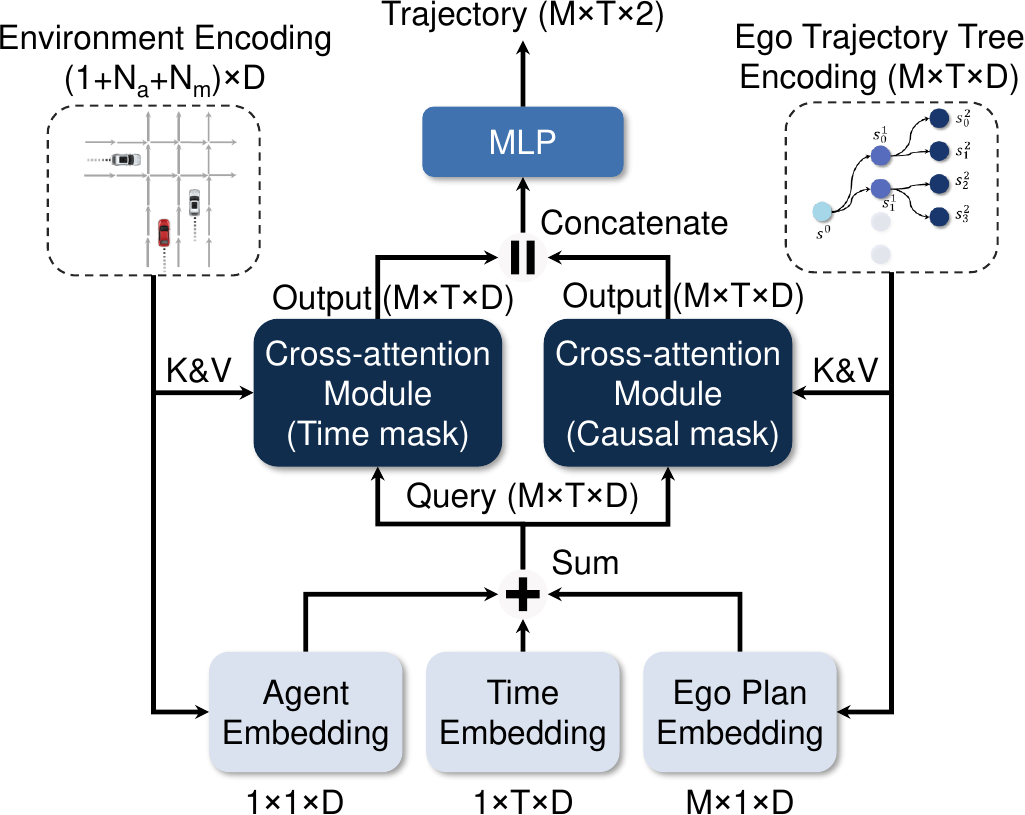}
    \caption{The detailed structure of the Transformer decoder for agent-wise trajectory prediction.}
    \label{fig:2}
    \vspace{-0.3cm}
\end{figure}

\cref{fig:2} illustrates the decoder for a single agent. It consists of two cross-attention Transformer modules that gather information from the environment context and the ego plan, respectively. The query input for the cross-attention module is derived from three sources: an agent history embedding directly retrieved from the corresponding position in the environment encoding, a learnable time embedding (to differentiate between time steps), and the ego plan embedding (target points of the ego agent that correspond to distinct branches). Multi-axis attention \cite{nayakanti2023wayformer} is applied to address modal and temporal dimensions in the query. 
Notably, we employ a time mask to suppress attention from invalid time steps or modalities and a casual mask to maintain the causal relationships \cite{chen2023tree} among modalities and timesteps. Finally, the outputs from the two attention modules are concatenated and processed through an MLP to decode the future trajectories of the agent. The decoding process is completed in a single shot because of the use of time embeddings. 
All agents share the same decoder. For different stages, the same decoder is employed to predict time-varying trajectories.

\begin{figure}[t]
    \centering
    \includegraphics[width=0.7\linewidth]{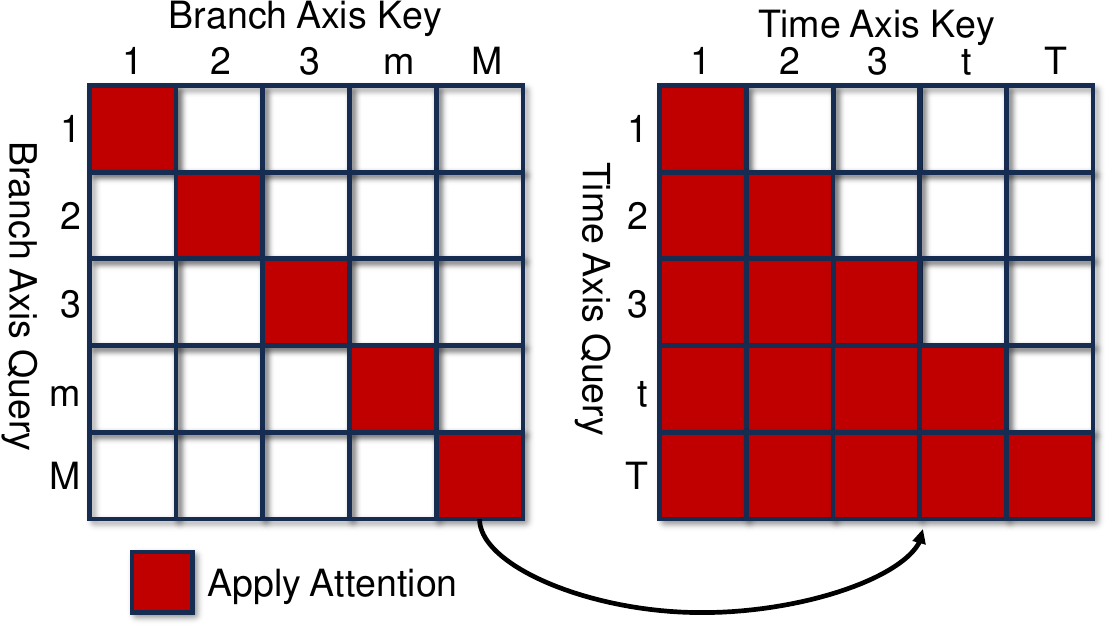}
    \caption{Illustration of casual masked attention for the ego trajectory tree.}
    \label{fig:3}
    \vspace{-0.3cm}
\end{figure}

\textbf{Casual masked attention}. 
\cref{fig:3} shows how we maintain causality in planning and conditional prediction using masked attention. For a given branch $m$, the causal mask excludes any information from other branches. Within each specific branch, predictions up to time step $t$ remain unaffected by any subsequent information beyond that time point. Notably, any invalid branches or time steps in the query or key/value tensors are masked during attention calculations. This masked attention mechanism can efficiently parallelize the predictions for all branches and timesteps in the scenario tree, while also preserving causality.

\subsection{Cost Function}
The cost of a candidate trajectory (branch) is the weighted sum of cost features $f^j$ over the time horizon:
\begin{equation}
c(s, p) = \frac{1}{t_F - t_0} \sum_{t=t_0}^{t_F} \sum_{j} \omega_j f^j (s(t), p(t)),
\end{equation}
where $s(t)$ is the ego state at time $t$, and $p(t)$ is the ego-conditional predicted state of other agents, and $\omega_j$ are learned context-aware weights.
The feature vectors $f^j, j=[e, i, c]$ consist of ego features $f^e$, learned interaction features $f^i$, and collision features $f^c$.  

The ego features $f^e$ consists of normalized ego jerk, acceleration, speed, and lateral acceleration. Through learned weights, ego features can balance comfort and progress.

The learned interaction features $f^i$ capture interaction between the ego and other agents. We employ an MLP to encode the relative attributes of the ego and other agents (relative position, heading, and velocity) into a higher-dimensional space (256), followed by max-pooling along the agent dimension. From this latent vector, we decode low-dimensional latent features (5) using another MLP:
\begin{equation}
f^i(s(t), p(t)) = \text{MLP} \left( \text{MaxPool} \left( \text{MLP} \left(p(t) - s(t) \right) \right) \right).
\end{equation}

The collision feature $f^c$ is computed with a handcrafted function that explicitly captures safety. Specifically, we apply a Gaussian radial basis function to the distances between the ego and other agents:
\begin{equation}
f^c(s(t), p(t)) = \sum_{i} \exp \left(  - 0.2 \parallel s(t) - p(i, t) \parallel^2 \right).
\end{equation} 

The cost features are summed with learned context-aware weights $\omega_j$. Specifically, they are decoded with an MLP from the ego’s encoding feature in the Transformer encoder model. The weights comprise 10 dimensions, encompassing ego (4), interaction (5), and collision (1) features.

Note that one could add additional cost terms, e.g., for lane keeping and route following. We did not find this necessary as our planner is constrained to follow a target lane through its trajectory generator (see~\cref{sec:planning}). 

\subsection{Model Training}
For training the conditional prediction task, we select the situation that most closely matches the ground-truth ego trajectory and then apply the smooth L1 loss to the predicted trajectories of the surrounding agents. 
\begin{equation}
\small
\mathcal{L}_{p} = \frac{1}{t_F - t_0} \sum_{t=t_0}^{t_F} \frac{1}{N_a} \sum_{i=1}^{N_a} \text{SmoothL1} \left( p^{*}(i,t) - p^{gt}(i,t) \right),   
\end{equation}
where $p^*$ is the prediction branch linked to the ego plan that is closest to the ground truth, and $p^{gt}$ represents the ground-truth trajectories of other agents.

Notably, our approach captures the multi-modality of the prediction outcomes by learning agent responses to different ego plans. Regarding the cost function, we employ maximum entropy IRL \cite{ziebart2008maximum}, which sets the probability of a plan to be $P(s_{m}) = \frac{e^{-c(s_m)}}{\sum_{n} e^{-c(s_n)}}$. Accordingly, we implement cross-entropy loss by comparing the softmax-normalized scores of each planned trajectory with the ground-truth label $y_m$. 
\begin{equation}
\mathcal{L}_{c} = -\sum_{m=1}^M y_{m} \log \left( P(s_{m}) \right).
\end{equation}
This approach enables the prediction-planning structure to be differentiable, directly linking agent predictions to overall planning performance. The total loss is then:
\begin{equation}
\mathcal{L} = \sum_{j=1}^{N_l}{\beta^j (\mathcal{L}^j_{p} + \alpha \mathcal{L}^j_{c})},
\end{equation}
where $\mathcal{L}^j_{p}$ and  $\mathcal{L}^j_{c}$ are losses for planning stage $j$; $\alpha=0.1$ is the weight assigned to the IRL loss; and $\beta^j$ is the weight associated with stage $j$. Given our priority on closed-loop planning, we assign a higher weight to the loss in the first stage, with $\beta^1=1.0$ and $\beta^2=0.2$. This setting enables the model to identify promising actions in the initial stage, which helps guide the tree expansion.

\section{Experiments}

\subsection{Experimental setup}
We experimentally verify our proposed method using the nuPlan dataset and its associated simulator \cite{nuplan}. The training and testing phases involve a set of labeled scenario types from the nuPlan competition, but we excluded certain static scenario types, resulting in 10 dynamic scenario types.
For training, we extract a total of 100k scenarios from the validation subset regarding the mentioned scenario types, and each scenario has a future time horizon of 8 seconds. For testing, we select 20 scenarios (each lasting 15 seconds) from the test subset for each scenario type, leading to a total of 200 scenarios.

Our evaluation covers both prediction and planning aspects. 
 The planning test encompasses three tasks: open-loop (OL), closed-loop with non-reactive agents (CL-NR), and closed-loop with reactive agents (CL-R). We report the average \emph{planning score} defined by the nuPlan simulator \cite{nuplan}. The score captures safety, efficiency, and comfort for CL-NR and CL-R, and similarity to human driving for OL. We also evaluate prediction metrics, specifically, average and final displacement error (ADE and FDE) for non-ego agents conditioned on the selected ego trajectory plan.

\subsection{Implementation Details}
The entire planning horizon spans $T=8$ seconds in $N_l=2$ stages with maximum $M=30$ branches. In the first stage (short-term, $t^1_0 = 0s$, $t^1_F = 3s$), we consider three reference paths and allow for a maximum of 30 sampled target states. After obtaining the prediction results, we retain only the top 5 nodes and proceed to expand each node with 6 target states for the second stage (long-term, $t^2_0 = 3s$, $t^2_F = 8s$). 


The model is trained using an NVIDIA RTX 3080 GPU. The batch size is 16 and the total number of training epochs is 20. We use an AdamW optimizer, and the learning rate is initialized at 1e-4 and reduced 50\% every 2 epochs after the 10th epoch. The model inference is conducted on the same GPU, while other computations run on an AMD 3900X CPU.

\subsection{Main Results}

\textbf{Closed-loop planning}.
\cref{tab1} presents the results of closed-loop planning tests. For a fair comparison, for the imitation learning-based baselines the output trajectories are projected onto the reference path. Our DTPP planner shows significantly better scores compared to learning-based policies, which often face causal confusion issues, as well as the rule-based IDM planner. Our method's performance closely matches that of the PDM(-Closed) method \cite{Dauner2023ARXIV}, which is the top-performing planning method on the leaderboard but is highly optimized for the nuPlan challenge metrics. Moreover, our method outperforms the baseline TPP method. For a detailed comparison with TPP, please refer to the following. 

\begin{table}[htp]
\caption{Closed-loop planning performance}
\centering
\resizebox{\linewidth}{!}{
\begin{tabular}{@{}l|cccc@{}}
\toprule
Method                              & NR Score ($\uparrow$)     & NR Collision ($\downarrow$)   & R Score  ($\uparrow$)     & R Collision ($\downarrow$) \\ \midrule
LaneGCN \cite{liang2020learning}    & 0.6356                    & 0.19                          & 0.6451                    & 0.185                               \\
Urban Driver \cite{scheel2022urban} & 0.6482                    & 0.18                          & 0.6598                    & 0.18                               \\
IDM \cite{albeaik2022limitations}   & 0.6396                    & 0.16                          & 0.6168                    & 0.15                            \\
TPP \cite{chen2023tree}             & 0.7388                    & 0.10                          & 0.7699                    & 0.065                               \\
PDM \cite{Dauner2023ARXIV}          & \textbf{0.9061}           & 0.035                         &\textbf{0.9150}            & \textbf{0.015}        \\
DTPP (Ours)                         &   0.8964                  & \textbf{0.025}                &  0.8978                   &  0.025                          \\ \bottomrule
\end{tabular}
}
\label{tab1}
\vspace{-0.2cm}
\end{table}

\textbf{Comparison with TPP}.
Our planner is based on the TPP method \cite{chen2023tree}, but it differs in two major ways: learnable cost function and node pruning. Firstly, cost learning is crucial to ensure accurate evaluation of actions and effective heuristics for guiding tree expansion. Secondly, pruning nodes can significantly reduce the number of branches in the scenario tree, which in turn improves the model's runtime. For planners without node pruning, we enlarge the size of the trajectory tree tensor to encompass all candidate trajectories. To make a fair comparison, the baseline TPP method employs the same conditional prediction model and trajectory generator as ours. 

The results of closed-loop planning tests and runtime analysis are shown in \cref{tab2}. The baseline TPP method, without cost learning, shows subpar performance in closed-loop planning. Node pruning can make performance much worse when cost learning is not included. However, with cost learning, even when done separately, planning performance significantly improves and pruning can be effective without affecting the planning performance. This is because, in the first stage, we can already choose the promising nodes and safely remove the other nodes. Additionally, joint training of prediction and cost evaluation models outperforms separate training, which can lead to suboptimal planning performance. Our DTPP method, which adopts joint learning of prediction and cost evaluation model, yields the best results. Using node pruning does not affect planning performance, but can significantly reduce runtime (by nearly half).

\begin{table}[htp]
\caption{Comparison with TPP on closed-loop planning}
\setlength{\tabcolsep}{3pt}
\centering
\resizebox{\linewidth}{!}{
\begin{tabular}{@{}l|ccccc@{}}
\toprule
                    & Cost Learning         & Pruning         & CL-NR Score      & CL-R Score       & Runtime (ms)\\ \midrule
TPP                 &  --                   &                 &  0.7388          &  0.7699          & 123.9\\
TPP w/ pruning      &  --                   &   \checkmark    &  0.3205          &  0.3228          & -- \\
                    & Separate              &                 &  0.8134          &  0.8106          &  -- \\
                    & Separate              &  \checkmark     &  0.7936          &  0.8185          & --\\  
DTPP w/o pruning    & Joint                 &                 &  0.8948          &  0.8933          &  199.6   \\
DTPP                & Joint                 &  \checkmark     &  \textbf{0.8964} & \textbf{0.8978}  & \textbf{98.0}   \\ \bottomrule
\end{tabular}
}
\label{tab2}
\vspace{-0.2cm}
\end{table}

\begin{figure}[htp]
\centering
\includegraphics[width=\linewidth]{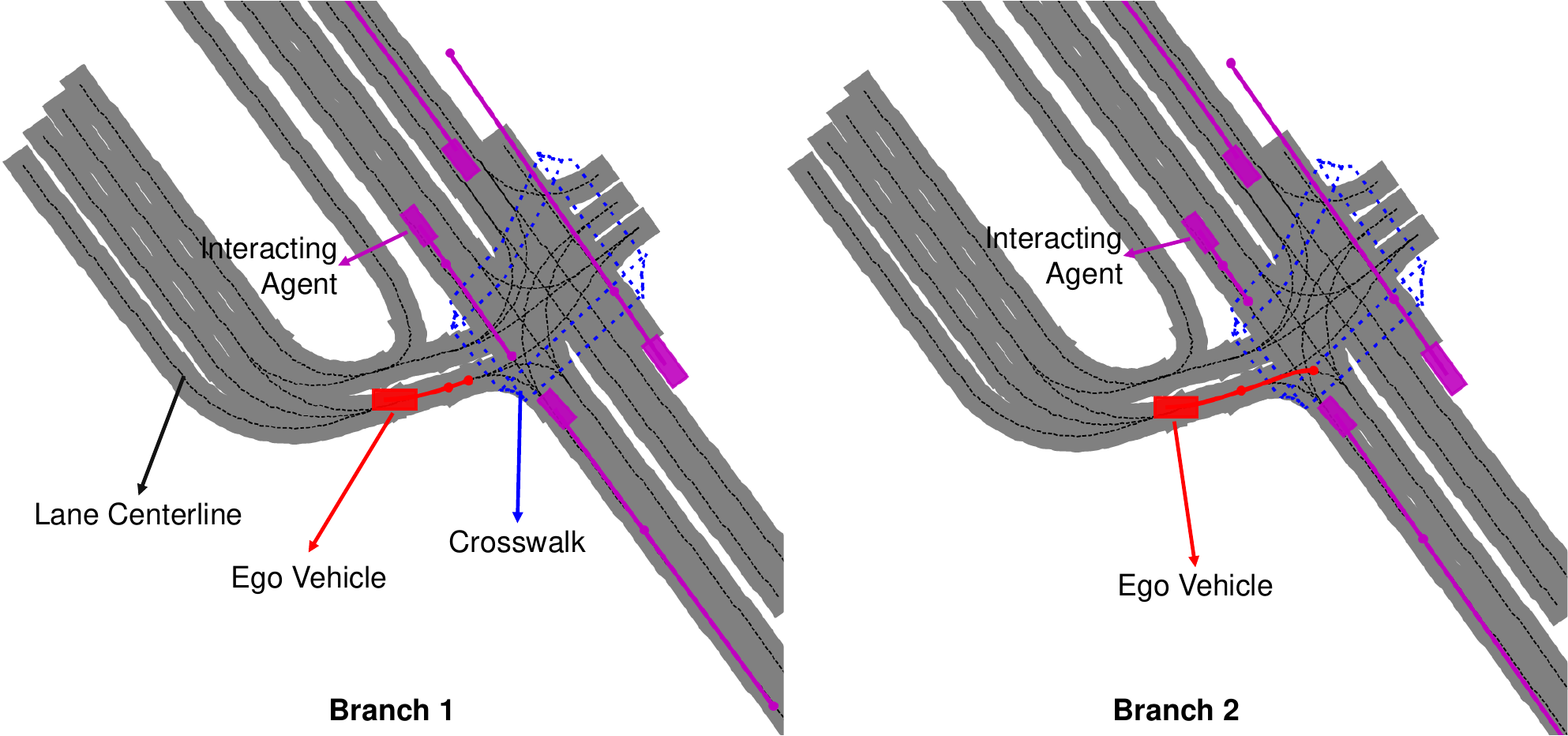}
\caption{Qualitative results of ego-conditioned prediction. Our model predicts reasonable behaviors for other agents given different ego plans.}
\label{fig4}
\vspace{-0.2cm}
\end{figure}

\textbf{Ego-conditioned prediction}. 
\cref{fig4} shows qualitative examples of ego-conditioned predictions at an intersection, where the ego vehicle intends to merge onto the main road. Given different plans for the ego vehicle, our model predicts distinct and reasonable responses from interacting agents. This feature helps the cost function model to evaluate situations more effectively and generate more accurate scores.

\textbf{Comparison of prediction models}. 
We compare the performance of our prediction model against two other baseline models, while keeping other parts of the planning framework unchanged. \textit{No ego-conditioned model (non-EC)}: information regarding the future trajectories of the ego vehicle is not incorporated into the model. \textit{Early fusion model (EF-EC)}: the ego vehicle’s future information is incorporated into the encoder part of the model and thus the model is queried iteratively for different plans. Note that the prediction metrics are derived from the branch selected by the cost evaluation model, and results from the non-EC method are directly employed for calculation. The results in \cref{tab4,tab5} indicate that ego-conditioned prediction models lag slightly in terms of accuracy compared to the non-EC model. However, they could enhance planning performance for both open-loop and closed-loop testing. Further, our proposed query-centric approach outperforms the early fusion method. 

\begin{table}[htp]
\caption{Model Comparison for Prediction Performance}
\centering
\begin{tabular}{@{}l|cccc@{}}
\toprule
Method       & ADE (3s)         & FDE (3s)         & ADE (8s)           & FDE (8s) \\ \midrule
non-EC       & \textbf{0.3365}  & \textbf{0.6593}  & \textbf{1.0515}    & \textbf{3.0171}   \\
EF-EC        & 0.3763           & 0.7296           & 1.2542             & 3.3602             \\
Ours         &  0.3636          & 0.6930           & 1.1797	            & 3.1237             \\ \bottomrule
\end{tabular}
\label{tab4}
\vspace{-0.2cm}
\end{table}

\begin{table}[htp]
\caption{Model Comparison for Planning Performance}
\centering
\begin{tabular}{@{}l|cccc@{}}
\toprule
Method       & OL Score         & CL-NR Score       & CL-R Score  \\ \midrule
non-EC       & 0.7533           & 0.8398            & 0.8411          \\
EF-EC        & 0.7718           & 0.8604            & 0.8655          \\
Ours         & \textbf{0.7911}  &  \textbf{0.8964}  & \textbf{0.8978}  \\ \bottomrule
\end{tabular}
\label{tab5}
\vspace{-0.2cm}
\end{table}

\cref{tab6} presents the runtime of planners with different prediction models. Our proposed ego-conditioned model can predict the outcomes of different branches in parallel, leading to significant enhancements in planning performance while incurring only a marginal increase in computation time than the non-EC model. Compared to the early fusion model, our method only needs to encode the scene once and avoids the need for iterative encoding of different branches.

\begin{table}[htp]
\centering
\caption{Runtime Analysis ($ms$) of Prediction models}
\begin{tabular}{@{}l|ccccc|c@{}}
\toprule
Method       & Encode  & Expand & Predict & Expand & Predict & Total \\ \midrule
Non-EC       & 6.8     &  8.2   & 31.0    &  14.8   & 29.8   & 90.6 \\
EF-EC        & --      & 8.2    & 300.8   &  14.8   & 275.5  & 599.3  \\
Ours         & 6.8     & 8.2    & 34.9    &  14.8   &  33.3  &  98.0  \\ \bottomrule
\end{tabular}
\label{tab6}
\vspace{-0.2cm}
\end{table}



\subsection{Ablation Studies}
\textbf{Influence of cost function learning}.
We investigate the influence of cost function and report the results in \cref{tab8}. The prediction metric and closed-loop planning score are used to evaluate the trained models with different cost-learning settings. The results indicate that the learned latent interaction feature plays a pivotal role in our model, ensuring the superior performance of our method in planning. The inclusion of collision potential in training, as well as the use of context-aware cost weights, proves advantageous for enhancing the final closed-loop planning performance. A noteworthy finding is that better prediction performance may not translate to better closed-loop planning, highlighting the importance of co-designing and joint training.

\begin{table}[htp]
\caption{Ablation results of the cost function}
\centering
\resizebox{\linewidth}{!}{
\begin{tabular}{@{}l|ccc@{}}
\toprule
Method                  & Prediction FDE (3s)    & CL-NR Score          & CL-R Score \\ \midrule
W/o latent interaction  & \textbf{0.5848}        & 0.7930               & 0.7968           \\
W/o collision potential & 0.6262                 & 0.8536               & 0.8650             \\
W/o variable weights    & 0.6483                 & 0.8611               & 0.8497            \\
Base (Ours)             & 0.6830                 & \textbf{0.8964}      & \textbf{0.8978}   \\ \bottomrule
\end{tabular}
}
\label{tab8}
\vspace{-0.2cm}
\end{table}

\textbf{Influence of tree-structured planning}.
In \cref{tab9}, we compare our DTPP approach with a single-stage trajectory planning method. We investigate different planning horizons for the single-stage planner and train the models using only single-stage loss. The results show that single-stage planners or models have better prediction accuracy. However, our multi-stage (tree-based) planning approach significantly outperforms the single-stage approach in closed-loop planning, whether the planning horizon is short (3 s) or long (8 s).

\begin{table}[htp]
\caption{Benefits of tree-structured planning}
\centering
\begin{tabular}{@{}l|ccc@{}}
\toprule
Method         & Prediction FDE (3s)   & CL-NR Score           & CL-R Score \\ \midrule
Single-3s      &   0.5788              &  0.7087               &  0.7044        \\
Single-8s      &  \textbf{0.5635}      &  0.7799               &  0.7558         \\
DTPP (Ours)    & 0.6830                & \textbf{0.8964}       & \textbf{0.8978}             \\ \bottomrule
\end{tabular}
\label{tab9}
\vspace{-0.2cm}
\end{table}

\section{CONCLUSIONS}

We propose DTPP, a differentiable joint learning framework for prediction and cost modeling, specifically designed for a tree policy planner. 
Our prediction model is a query-centric Transformer network with efficient ego conditioning. The cost model combines learned and handcrafted features with learned context-aware weights.
The experimental results for planning and prediction on real-world driving data show that our prediction model yields significantly better performance and efficiency. Joint training is crucial for achieving optimal planning performance, and our tree-based planner significantly outperforms single-stage trajectory planning.
Future work may extend our framework to multi-modal probabilistic predictions and perform real-world experiments.




\bibliographystyle{ieeetr} 
\bibliography{IEEEexample}

\end{document}